\documentclass[letterpaper, 10 pt, conference]{ieeeconf}  % Comment this line out if you need a4paper

\IEEEoverridecommandlockouts                              % This command is only needed if 
\usepackage{array}
\usepackage{amsmath}
\usepackage{amssymb}
\usepackage{graphicx}
\usepackage{color}
\usepackage{threeparttable}
\usepackage{multirow}
\usepackage{makecell}
\usepackage[hidelinks]{hyperref}

\usepackage{soul}
\title{\LARGE \bf
A Unified Benchmark for RCM-Constrained Visual Servoing: Modeling–Controller Interaction and Robustness Analysis in Laparoscopic Robots
}

\author{Jing Zhang, Mengtang Li, \textit{Member, IEEE}% <-this % stops a space
\thanks{All authors are with School of Intelligent Systems Engineering, Shenzhen Campus of Sun Yat-sen University, Shenzhen, China.
\textit{Corresponding author: Mengtang Li (e-mail: limt29@mail.sysu.edu.cn).}}
}

\begin{document}

\maketitle
\thispagestyle{empty}
\pagestyle{empty}

%%%%%%%%%%%%%%%%%%%%%%%%%%%%%%%%%%%%%%%%%%%%%%%%%%%%%%%%%%%%%%%%%%%%%%%%%%%%%%%%
\begin{abstract}
	In robot-assisted laparoscopic minimally invasive surgery (MIS), 
    accurate enforcement of the remote center of motion (RCM) constraint is critical for safe and stable automatic field-of-view (FoV) adjustment. 
	Although control-based RCM strategies are widely adopted due to their flexibility and cost-effectiveness, 
    systematic comparison of different RCM formulations and image-based visual servoing (IBVS) frameworks remains challenging due to the lack of a unified and reproducible benchmark. 
	This paper presents an open-source simulation framework integrating three representative RCM modeling approaches and six IBVS-based control architectures within a unified velocity-level formulation, 
    enabling controlled and consistent evaluation. 
	Through structured case studies, the framework reveals key structural sensitivities arising from modeling and controller interactions, 
    including the impact of tangent-plane definition, constraint dimensionality, open- versus closed-loop enforcement, and robustness near kinematic singularities. 
	All resources are released and demostrations are provided in the supplementary video, 
    providing a reproducible foundation for RCM-constrained visual servoing research.
\end{abstract}

%%%%%%%%%%%%%%%%%%%%%%%%%%%%%%%%%%% Introduction %%%%%%%%%%%%%%%%%%%%%%%%%%%%%%%%%%%%%%%%%%%%%%%
\section{Introduction}
Surgical robots have been widely adopted in various medical applications over the past decades due to their high precision and intelligence~\cite{2025ciutiRobotic}.
In robot-assisted laparoscopic surgery, robots can manipulate the laparoscope to reduce surgeon workload and improve procedural stability~\cite{2025pereiraInnovative}.
Recent advances in visual servoing~\cite{2023fozilovEndoscope,2025zhouLaparoscope} and learning-based perception~\cite{2019rivas-blancoTransferring,2022liLearning} further enable automatic field-of-view (FoV) adjustment during surgery.
A fundamental requirement in these systems is the enforcement of the remote center of motion (RCM) constraint at the trocar, 
which restricts the laparoscope to pivot about a fixed incision point, thereby preventing tissue damage~\cite{2024zhangState}.

The RCM methodologies can be broadly categorized into design-based and control-based strategies~\cite{2024zhangState}.
Design-based approaches rely on specialized mechanical structures to physically enforce the pivot constraint, 
as exemplified by commercial systems such as the da Vinci platform.
Although mechanically precise, these solutions are often complex and costly~\cite{2025wangDynamic}.
In contrast, control-based RCM strategies enforce the constraint at the kinematic level and can be implemented on general-purpose robotic manipulators without dedicated mechanisms.
This flexibility has led to a variety of RCM modeling and control formulations in the literature~\cite{2010azimianConstrained,2017sandovalNew,2013aghakhaniTask}.

Existing control-based RCM formulations can be divided into two main categories.
Calculation-based approaches determine the RCM point through geometric relationships, 
such as tangent-plane constraints~\cite{2010azimianConstrained} or projection onto the laparoscope axis~\cite{2017sandovalNew}.
These models have been combined with control architectures like
quadratic programming (QP) schemes~\cite{2025zhouControl,2023colanConstrained}.
Variable-based approaches, on the other hand, introduce a virtual joint variable to parameterize the RCM position along the shaft~\cite{2013aghakhaniTask}, 
which has been adopted in both rigid and flexible laparoscopic systems~\cite{2025huangAccelerated,2025caoUncalibrated}.
Table~\ref{tab:literature} summarizes representative FoV-adjustment works using control-based RCM methods.

\begin{table}
    \centering
    \begin{threeparttable}
    \caption{Summary of automatic FoV adjustment works with control-based RCM methods.}
    \label{tab:literature}
    \begin{tabular}{cccc}
        \hline
        
        \hline
        RCM type & Work  & Robot type & Controller\tnote{1} \\
        \hline
        \multirow{3}{*}{\makecell{Calculation\\based}} 
        & Fozilov et al., 2023~\cite{2023fozilovEndoscope} & Rigid & QP \\
        & Zhou et al., 2025~\cite{2025zhouControl} & Rigid & QP \\
        & Yang et al., 2026~\cite{2026yangMultiinstrument} & Rigid  & IK \\
        \hline
        \multirow{3}{*}{\makecell{Variable\\based}} 
        & Aghakhani et al., 2013~\cite{2013aghakhaniTask} & Rigid & PI \\
        & Huang et al., 2025~\cite{2025huangAccelerated} & Flexible & QP \\
        & Cao et al., 2025~\cite{2025caoUncalibrated} & Rigid  & QP \\
        
        \hline

        \hline
        
    \end{tabular}
    \begin{tablenotes}
        \footnotesize
        \item[1] IK: Inverse kinematics-based; QP: Quadratic programming-based; PI: Pseudoinverse-based.
    \end{tablenotes}
    \end{threeparttable}
\end{table}

While prior studies have analyzed the properties of different RCM generation algorithms~\cite{2017marinhoComparison,2022liComparisons}, 
existing comparisons neglect several issues in practical applications:
\begin{enumerate}
	\item How different RCM modeling choices behave under identical image-based visual servoing (IBVS) tasks?
	\item How controller architecture interacts with RCM constraint formulation?
	\item What practical sensitivities arise from seemingly minor modeling decisions?
\end{enumerate}

Moreover, publicly available implementations that allow reproducible and controlled comparison of these methods are still lacking. 
To address this gap, this paper develops a unified and open-source simulation infrastructure that integrates three representative RCM modeling methods and six IBVS-compatible control frameworks within a consistent velocity-level formulation. 
Rather than proposing a new RCM method, the objective of this paper is to provide a structured benchmark that enables systematic evaluation of modeling and controller interactions under identical conditions. 
Through a series of carefully designed simulation cases, the framework exposes practical sensitivities related to tangent-plane definition, constraint dimensionality, open- versus closed-loop enforcement, and robustness near kinematic singularities. 
The contributions of this paper are summarized as follows:
\begin{enumerate}	
	\item A unified velocity-level formulation that expresses three representative RCM models within a common constraint structure.	
	\item A modular benchmark integrating three RCM models and six control frameworks for controlled and reproducible comparison.	
	\item A systematic study exposing structural sensitivities between modeling choices and controller architectures, 
    including constraint dimensionality and singularity robustness.
	\item An open-source playground providing a reproducible baseline for RCM-constrained visual servoing research.
\end{enumerate}

The remainder of this paper is organized as follows:
Section~II introduces the modeling of the RCM constraint.
Section~III presents the visual servoing task.
Section~IV describes the different control frameworks.
Section~V shows the simulation results, and Section~VI makes further discussions.
Finally, Section~VII concludes the paper.

%%%%%%%%%%%%%%%%%%%%%%%%%%%%%%%%%%%%%%%%%%%%%%%%%%%%%%%%%%%%%%%%%%%%%%%%%%%%%%%%
\section{Modeling of the RCM constraint}
\begin{figure}
    \centering
    \includegraphics[width=0.45\textwidth]{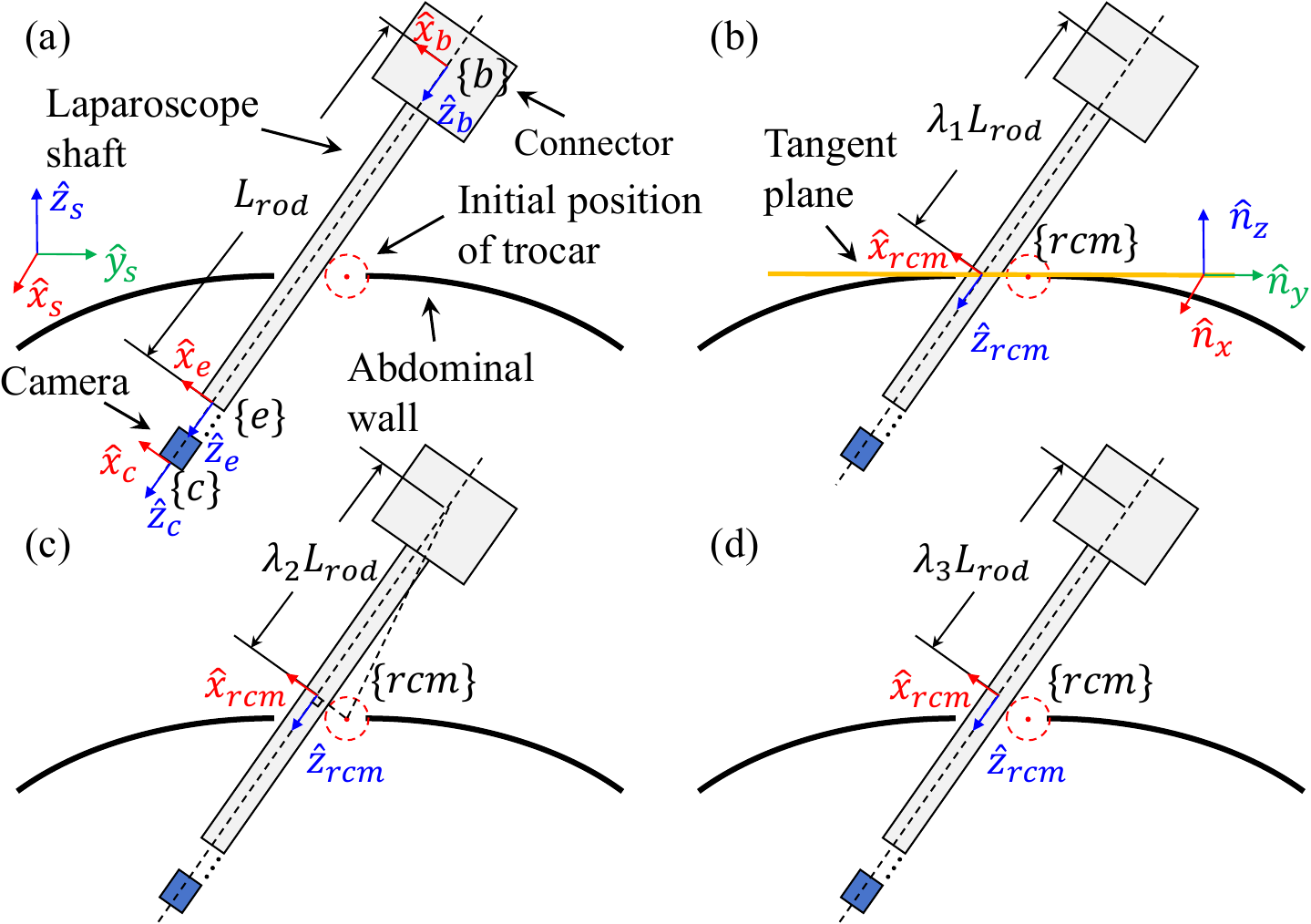}
    \caption{Schematic diagrams of RCM modeling methods.
    (a) Coordinate frames involved.
    (b) RCM 1~\cite{2010azimianConstrained}.
    (c) RCM 2~\cite{2017sandovalNew}.
    (d) RCM 3~\cite{2013aghakhaniTask}.}
    \label{fig:coordinates}
\end{figure}

This section introduces three representative RCM modeling methods.
Fig.~\ref{fig:coordinates} illustrates the coordinate frames involved. 
Frame $\{s\}$ denotes the world frame in simulation or the robot base frame in practice.
The camera frame $\{c\}$ is located at the laparoscope tip, with its $z$-axis aligned with the optical axis and its $x$-axis pointing to the right of the image. 
Frames $\{e\}$, $\{b\}$, and $\{rcm\}$ are defined at the distal end of the laparoscope shaft, the connector, and the RCM point, respectively. 
These frames share the same initial orientation as $\{c\}$ and remain aligned during motion. 
For rigid laparoscopes, $\{c\}$ and $\{e\}$ coincide, whereas for flexible systems a bending section exists between them. 
Figs.~\ref{fig:coordinates}(b)-(d) depict the three RCM modeling approaches. 
In all cases, the RCM position along the shaft is parameterized by a scalar $\lambda \in (0,1)$, 
defined as the ratio $L_{out}/L_{rod}$, where $L_{out}$ is the distance from $\{b\}$ to $\{rcm\}$
and $L_{rod}$ is the shaft length from $\{b\}$ to $\{e\}$. 
The scalar is denoted as $\lambda_i$ for the $i$-th method.

% ---------------------------------------------------------------------------------------------- %
\subsubsection{RCM 1~\cite{2010azimianConstrained}}
As shown in Fig.~\ref{fig:coordinates}(b), this method utilizes a tangent plane at the trocar point to calculate the scalar $\lambda_{1}$:
\begin{equation}
    \lambda_{1} = \frac{\hat{\boldsymbol{n}}_{z}^{T}({}^{s}\boldsymbol{p}_{b}-{}^{s}\boldsymbol{p}_{tro})}
    {\hat{\boldsymbol{n}}_{z}^{T}({}^{s}\boldsymbol{p}_{b}-{}^{s}\boldsymbol{p}_{e})}
\end{equation}
where ${}^{s}\boldsymbol{p}_{b} \in \mathbb{R}^{3}$ and ${}^{s}\boldsymbol{p}_{e}\in \mathbb{R}^{3}$ 
are the positions of frame $\{b\}$ and frame $\{e\}$ in frame $\{s\}$, respectively,
${}^{s}\boldsymbol{p}_{tro}\in \mathbb{R}^{3}$ is the initial position of the trocar point in frame $\{s\}$,
and $\hat{\boldsymbol{n}}_{z}\in \mathbb{R}^{3}$ is the unit vector normal to the tangent plane of the trocar point:
\begin{equation}
    \hat{\boldsymbol{n}}_{z}^{T}\boldsymbol{p}+d=0
    \label{eq:trocar_plane}
\end{equation}
where $\boldsymbol{p}\in\mathbb{R}^{3}$ is the position of an arbitrary point on the tangent plane,
and $d$ is a constant, which is calculated by substituting $\boldsymbol{p}={}^{s}\boldsymbol{p}_{tro}$ into Eq.~(\ref{eq:trocar_plane}).
Furthermore, the position ${}^{s}\boldsymbol{p}_{rcm}\in \mathbb{R}^{3}$ and linear velocity ${}^{s}\boldsymbol{v}_{rcm}\in \mathbb{R}^{3}$ of the RCM point in frame $\{s\}$ is:
\begin{align}
    {}^{s}\boldsymbol{p}_{rcm}&=\lambda_{1}{}^{s}\boldsymbol{p}_{e}+(1-\lambda_{1}){}^{s}\boldsymbol{p}_{b} \\
    {}^{s}\boldsymbol{v}_{rcm}&=\lambda_{1}{}^{s}\boldsymbol{v}_{e}+(1-\lambda_{1}){}^{s}\boldsymbol{v}_{b}
    \label{eq:rcm1_velocity}
\end{align}
where ${}^{s}\boldsymbol{v}_{e}\in \mathbb{R}^{3}$ and ${}^{s}\boldsymbol{v}_{b}\in \mathbb{R}^{3}$
are the linear velocities of the distal end of the laparoscope shaft and the connector in frame $\{s\}$, respectively.
To ensure the RCM point does not have any lateral deflection in the tangent plane, the following constraint should be satisfied:
\begin{equation}
    \begin{bmatrix}
        \hat{\boldsymbol{n}}_{x}^{T};
        \hat{\boldsymbol{n}}_{y}^{T}
    \end{bmatrix}
    {}^{s}\boldsymbol{v}_{rcm} = \boldsymbol{0}_{2\times1}
    \label{eq:rcm1_constraint}
\end{equation}
where $\hat{\boldsymbol{n}}_{x}\in \mathbb{R}^{3}$ and $\hat{\boldsymbol{n}}_{y}\in \mathbb{R}^{3}$ are unit vectors spanning the tangent plane defined in Eq.~(\ref{eq:trocar_plane}).
% The selection of the tangent plane will be discussed in the simulation study.
Substitute Eq.~(\ref{eq:rcm1_velocity}) to Eq.~(\ref{eq:rcm1_constraint}):
\begin{equation}
    \boldsymbol{J}_{rcm,1}\dot{\boldsymbol{\Theta}}  = \boldsymbol{0}_{2\times1}
\end{equation}
\begin{equation}
    \boldsymbol{J}_{rcm,1}
     =\begin{bmatrix}
        \hat{\boldsymbol{n}}_{x}^{T};
        \hat{\boldsymbol{n}}_{y}^{T}
    \end{bmatrix}
    (\lambda_{1}{}^{s}\boldsymbol{J}_{e}+(1-\lambda_{1}){}^{s}\boldsymbol{J}_{b})
\end{equation}
where $\dot{\boldsymbol{\Theta}}\in \mathbb{R}^{N_{\Theta}}$ is the joint velocity and $N_{\Theta}$ is the number of joints of the laparoscopic robot,
and ${}^{s}\boldsymbol{J}_{e} \in \mathbb{R}^{3 \times N_{\Theta}}$ and ${}^{s}\boldsymbol{J}_{b} \in \mathbb{R}^{3 \times N_{\Theta}}$ 
are the Jacobian matrices mapping $\dot{\boldsymbol{\Theta}}$ to ${}^{s}\boldsymbol{v}_{e}$ and ${}^{s}\boldsymbol{v}_{b}$, respectively.

% ---------------------------------------------------------------------------------------------- %
\subsubsection{RCM 2~\cite{2017sandovalNew}}
As shown in Fig.~\ref{fig:coordinates}(c), this method projects the trocar point onto the axis of the laparoscope shaft to obtain $\lambda_{2}$:
\begin{equation}
    \lambda_{2} = {}^{s}\hat{\boldsymbol{z}}_{b}^{T}({}^{s}\boldsymbol{p}_{tro}-{}^{s}\boldsymbol{p}_{b})/L_{rod}
\end{equation}
where ${}^{s}\hat{\boldsymbol{z}}_{b}={}^{s}\boldsymbol{R}_{b}[0;0;1]$ is the unit vector along the $z$-axis of frame $\{b\}$ in frame $\{s\}$,
which coincides with the axis of the laparoscope shaft,
and ${}^{s}\boldsymbol{R}_{b} \in SO(3)$ is the rotation matrix of frame $\{b\}$ in frame $\{s\}$.
Based on this, the position and linear velocity of the RCM point are:
\begin{align}
    {}^{s}\boldsymbol{p}_{rcm} &={}^{s}\boldsymbol{p}_{b}+\lambda_{2}L_{rod}{}^{s}\hat{\boldsymbol{z}}_{b}
    \label{eq:rcm2_pos} \\
    {}^{s}\boldsymbol{v}_{rcm} &={}^{s}\boldsymbol{v}_{b}+{}^{s}\boldsymbol{\omega}_{b} \times (\lambda_{2}L_{rod}{}^{s}\hat{\boldsymbol{z}}_{b})
    \label{eq:rcm2_vel}
\end{align}
where ${}^{s}\boldsymbol{\omega}_{b}\in\mathbb{R}^{3}$ is the angular velocity of the connector in frame $\{s\}$.
Additionally, the position of the RCM point can also be extracted via ${}^{s}\boldsymbol{p}_{rcm}  = {}^{s}\boldsymbol{T}_{rcm}(1:3,4)$,
where ${}^{s}\boldsymbol{T}_{rcm} \in SE(3)$ is the homogeneous transformation matrix of frame $\{rcm\}$ in frame $\{s\}$,
which is calculated by:
\begin{equation}
    {}^{s}\boldsymbol{T}_{rcm} = {}^{s}\boldsymbol{T}_{b} \cdot \text{transl\_z}(\lambda_{2}L_{rod})
\end{equation}
where ${}^{s}\boldsymbol{T}_{b} \in SE(3)$ is the homogeneous transformation matrix of frame $\{b\}$ in frame $\{s\}$,
and $\text{transl\_z}(\cdot)$ is a function that generates a homogeneous transformation matrix representing a translation along the $z$-axis.
Furthermore, the velocity twist ${}^{rcm}\boldsymbol{\mathcal{V}}_{rcm}=[{}^{rcm}\boldsymbol{\omega}_{rcm};{}^{rcm}\boldsymbol{v}_{rcm}]$ of the RCM point
can be obtained from that of the connector ${}^{b}\boldsymbol{\mathcal{V}}_{b}=[{}^{b}\boldsymbol{\omega}_{b};{}^{b}\boldsymbol{v}_{b}]$ through the adjoint representation:
% \begin{equation}
%     \begin{bmatrix}
%         {}^{rcm}\boldsymbol{\omega}_{rcm} \\
%         {}^{rcm}\boldsymbol{v}_{rcm}
%     \end{bmatrix} = 
%     \begin{bmatrix}
%         {}^{rcm}\boldsymbol{R}_{b} & \boldsymbol{0}_{3\times3} \\
%         [{}^{rcm}\boldsymbol{p}_{b}]_{\times}{}^{rcm}\boldsymbol{R}_{b} & {}^{rcm}\boldsymbol{R}_{b}
%     \end{bmatrix}
%     \begin{bmatrix}
%         {}^{b}\boldsymbol{\omega}_{b} \\
%         {}^{b}\boldsymbol{v}_{b}
%     \end{bmatrix} 
%     \label{eq:rcm2_adj}
% \end{equation}
\begin{equation}
    \begin{bmatrix}
        {}^{rcm}\boldsymbol{\omega}_{rcm} \\
        {}^{rcm}\boldsymbol{v}_{rcm}
    \end{bmatrix} = 
    \begin{bmatrix}
        \boldsymbol{I}_{3\times3} & \boldsymbol{0}_{3\times3} \\
        [{}^{rcm}\boldsymbol{p}_{b}]_{\times} & \boldsymbol{I}_{3\times3}
    \end{bmatrix}
    \begin{bmatrix}
        {}^{b}\boldsymbol{\omega}_{b} \\
        {}^{b}\boldsymbol{v}_{b}
    \end{bmatrix} 
    \label{eq:rcm2_adj}
\end{equation}
where ${}^{rcm}\boldsymbol{\omega}_{rcm}\in\mathbb{R}^{3}$ and ${}^{rcm}\boldsymbol{v}_{rcm}\in\mathbb{R}^{3}$
(${}^{b}\boldsymbol{\omega}_{b}\in\mathbb{R}^{3}$ and ${}^{b}\boldsymbol{v}_{b}\in\mathbb{R}^{3}$) 
are the angular and linear velocities of the RCM point (the connector) in frame $\{rcm\}$ (frame $\{b\}$), respectively,
$\boldsymbol{I}_{3\times3}$ and $\boldsymbol{0}_{3\times3}$ denote the $3 \times 3$  identity and zero matrices, respectively,
${}^{rcm}\boldsymbol{p}_{b} = [0;0;-L_{out}]$ is the position of frame $\{b\}$ in frame $\{rcm\}$,
and $[\cdot]_{\times}$ is the skew-symmetric matrix operator.

To satisfy the RCM constraint, the linear velocities of the RCM point at $x$ and $y$ directions in frame $\{rcm\}$ should be controlled:
\begin{equation}
    \boldsymbol{J}_{rcm,2}(1:2,:)\dot{\boldsymbol{\Theta}} = {}^{rcm}\boldsymbol{v}_{rcm,des}
\end{equation}
\begin{equation}
    \boldsymbol{J}_{rcm,2}
    = \begin{bmatrix}
        [{}^{rcm}\boldsymbol{p}_{b}]_{\times} & \boldsymbol{I}_{3\times3}
    \end{bmatrix} 
    {}^{b}\boldsymbol{J}_{b}
\end{equation}
where ${}^{b}\boldsymbol{J}_{b} \in \mathbb{R}^{6 \times N_{\Theta}}$ 
is the Jacobian matrix mapping $\boldsymbol{\dot{\Theta}}$ to ${}^{b}\boldsymbol{\mathcal{V}}_{b}$,
and ${}^{rcm}\boldsymbol{v}_{rcm,des}\in \mathbb{R}^{2}$ contains the desired linear velocities of the RCM point at $x$ and $y$ directions in frame $\{rcm\}$, 
which is a zero vector for open-loop control of the position of the RCM point,
or a vector calculated based on the RCM error, defined as ${}^{s}\boldsymbol{p}_{tro} - {}^{s}\boldsymbol{p}_{rcm}$,
for closed-loop control.

% ---------------------------------------------------------------------------------------------- %
\subsubsection{RCM 3~\cite{2013aghakhaniTask}}
As shown in Fig.~\ref{fig:coordinates}(d), this method introduces $\lambda_{3}$ as a virtual joint variable
to derive the position and the linear velocity of the RCM point in frame $\{s\}$:
\begin{equation}
    {}^{s}\boldsymbol{p}_{rcm} = {}^{s}\boldsymbol{p}_{b}+\lambda_{3}({}^{s}\boldsymbol{p}_{e}-{}^{s}\boldsymbol{p}_{b})
\end{equation}
\begin{equation}
    {}^{s}\boldsymbol{v}_{rcm}={}^{s}\boldsymbol{v}_{b}
    +\dot{\lambda}_{3}({}^{s}\boldsymbol{p}_{e}-{}^{s}\boldsymbol{p}_{b})
    +\lambda_{3}({}^{s}\boldsymbol{v}_{e}-{}^{s}\boldsymbol{v}_{b})
\end{equation}
where $\dot{\lambda}_{3}$ is the velocity of the virtual joint variable $\lambda_{3}$.
It is worth noting that there is another equivalent way~\cite{2019sadeghianConstrained} to express the linear velocity of the RCM point:
\begin{equation}
    {}^{s}\boldsymbol{v}_{rcm} = {}^{s}\boldsymbol{v}_{b} + {}^{s}\boldsymbol{v}_{rcm}^{b}
    - [{}^{s}\boldsymbol{p}_{rcm}^{b}]_{\times} {}^{s}\boldsymbol{\omega}_{b}
\end{equation}
where ${}^{s}\boldsymbol{v}_{rcm}^{b}$ is the linear velocity of the RCM point related to frame $\{b\}$ in frame $\{s\}$,
and ${}^{s}\boldsymbol{p}_{rcm}^{b} = {}^{s}\boldsymbol{p}_{rcm} - {}^{s}\boldsymbol{p}_{b}$.
In this method, the RCM constraint equation is expressed as:
\begin{equation}
    \boldsymbol{J}_{rcm,3}    
    \begin{bmatrix}
        \dot{\boldsymbol{\Theta}};
        \dot{\lambda}_{3}
    \end{bmatrix} = {}^{s}\boldsymbol{v}_{rcm,des}
    \label{eq:rcm3_constraint}
\end{equation}
\begin{equation}
    \boldsymbol{J}_{rcm,3}
    = 
    \begin{bmatrix}
        {}^{s}\boldsymbol{J}_{b}+\lambda_{3}({}^{s}\boldsymbol{J}_{e}-{}^{s}\boldsymbol{J}_{b}) & {}^{s}\boldsymbol{p}_{e}-{}^{s}\boldsymbol{p}_{b}
    \end{bmatrix}
\end{equation}
where ${}^{s}\boldsymbol{v}_{rcm,des}\in \mathbb{R}^{3}$ is the desired linear velocity of the RCM point in frame $\{s\}$,
which holds the same assignment logic as ${}^{rcm}\boldsymbol{v}_{rcm,des}$ in RCM 2.
Eq.~(\ref{eq:rcm3_constraint}) can be changed to express under frame $\{rcm\}$ by left multiplying the rotation matrix ${}^{rcm}\boldsymbol{R}_{s}\in SO(3)$,
which describes the rotation of frame $\{s\}$ in frame $\{rcm\}$, on both sides.

% ---------------------------------------------------------------------------------------------- %
\subsubsection{Unified Expression}
In summary, the RCM constraint can be uniformly expressed as:
\begin{equation}
    \boldsymbol{J}_{rcm,i}\boldsymbol{\dot{u}} = \boldsymbol{v}_{rcm,des}
    \label{eq:rcm_constraint}
\end{equation}
where $\boldsymbol{u} = \boldsymbol{\Theta}$ and $\boldsymbol{\dot{u}} = \dot{\boldsymbol{\Theta}}$ for RCM 1 and RCM 2,
whereas for RCM 3, $\boldsymbol{u} = [\boldsymbol{\Theta}; \lambda_{3}]$ 
and $\boldsymbol{\dot{u}} = [\dot{\boldsymbol{\Theta}}; \dot{\lambda}_{3}]$,
$\boldsymbol{v}_{rcm,des} \in \mathbb{R}^{N_{rcm}}$ is the desired velocity of the RCM point,
and $N_{rcm}$ is the number of RCM constraint equations.

%%%%%%%%%%%%%%%%%%%%%%%%%%%%%%%%%%%%%%%%%%%%%%%%%%%%%%%%%%%%%%%%%%%%%%%%%%%%%%%%
\section{Visual servoing task}
This section formulates the visual servoing task for automatic FoV adjustment of laparoscopic robots.
Generally, the relationship between visual feature velocity and robot joint velocity is given by~\cite{2023huang4DOF}:
\begin{equation}
    \dot{\boldsymbol{s}} = \boldsymbol{J}_{t}\dot{\boldsymbol{\Theta}}
\end{equation}
where $\boldsymbol{s} \in \mathbb{R}^{N_{f}}$ is the visual feature, $\dot{\boldsymbol{s}}$ is the velocity of the visual feature,
$N_{f}$ is the number of visual features,
and $\boldsymbol{J}_{t}\in \mathbb{R}^{N_{f} \times N_{\Theta}}$ is the Jacobian matrix of the whole laparoscopic robot system.

The objective of visual servoing is to drive $\dot{\boldsymbol{s}}$ to a desired velocity $\dot{\boldsymbol{s}}_{des}$, computed from the feature tracking error.
Accordingly, the task is written as:
\begin{equation}
    \boldsymbol{J}_{t,i}\boldsymbol{\dot{u}} = \dot{\boldsymbol{s}}_{des}
    \label{eq:visual_servoing}
\end{equation}
where $\boldsymbol{J}_{t,i} = \boldsymbol{J}_{t}$ for $i=1,2$ and $\boldsymbol{J}_{t,3} = [\boldsymbol{J}_{t}, \boldsymbol{0}_{N_{f}\times1}]$.

%%%%%%%%%%%%%%%%%%%%%%%%%%%%%%%%%%%%%%%%%%%%%%%%%%%%%%%%%%%%%%%%%%%%%%%%%%%%%%%%
\section{Control frameworks}
This section introduces six different control frameworks for achieving the visual servoing task under the RCM constraint.
Specifically, two pseudoinverse (PI)-based control frameworks, three QP-based methods, and one inverse-kinematics (IK)-based approach are introduced.

% ---------------------------------------------------------------------------------------------- %
\subsection{PI-based Method}
PI-based methods calculate the joint velocity by directly solving Eq.~(\ref{eq:rcm_constraint}) and Eq.~(\ref{eq:visual_servoing}),
which represent the RCM constraint and visual servoing task, respectively.

\subsubsection{PI 1}
This PI-based method treats both RCM constraint and visual servoing task equally.
Combine Eq.~(\ref{eq:rcm_constraint}) and Eq.~(\ref{eq:visual_servoing}) to obtain the control value~\cite{2013aghakhaniTask}:
\begin{equation}
    \boldsymbol{\dot{u}} = 
    \begin{bmatrix}
        \boldsymbol{J}_{t,i};
        \boldsymbol{J}_{rcm,i}
    \end{bmatrix}^{\dagger}
    \begin{bmatrix}
        \dot{\boldsymbol{s}}_{des};
        \boldsymbol{v}_{rcm,des}
    \end{bmatrix} 
\end{equation}
where [$\cdot$]$^{\dagger}$ is the pseudoinverse operator.

% ---------------------------------------------------------------------------------------------- %
\subsubsection{PI 2}
In this method, the null-space projection is introduced to prioritize the RCM constraint over the visual servoing task,
which is similar to~\cite{2010azimianConstrained}:
\begin{equation}
    \boldsymbol{\dot{u}} = \boldsymbol{\dot{u}}_{rcm} + \boldsymbol{N}_{rcm,i}\boldsymbol{\dot{u}}_{t}
\end{equation}
\begin{equation}
    \boldsymbol{\dot{u}}_{rcm} = \boldsymbol{J}_{rcm,i}^{\dagger}\boldsymbol{v}_{rcm,des}
\end{equation}
\begin{equation}
    \boldsymbol{\dot{u}}_{t} = (\boldsymbol{J}_{t,i}\boldsymbol{N}_{rcm,i})^{\dagger}(\dot{\boldsymbol{s}}_{des}-\boldsymbol{J}_{t,i}\boldsymbol{\dot{u}}_{rcm})
\end{equation}
where $\boldsymbol{N}_{rcm,i} = \boldsymbol{I}-\boldsymbol{J}_{rcm,i}^{\dagger}\boldsymbol{J}_{rcm,i}$ is the null space projector of $\boldsymbol{J}_{rcm,i}$,
and $\boldsymbol{I}$ is the identity matrix of appropriate size.

% ---------------------------------------------------------------------------------------------- %
\subsection{QP-based Method}
In practical applications, laparoscopic robots are subject to physical constraints such as joint angle and velocity limits, 
which are difficult to incorporate directly into PI-based methods.
QP-based frameworks address this limitation by enforcing the RCM constraint and visual servoing task while explicitly incorporating these physical bounds as inequality constraints:

\begin{equation}
    [\boldsymbol{u}_{min}; \boldsymbol{\dot{u}}_{min}] 
    \leq [\boldsymbol{u}; \boldsymbol{\dot{u}}] 
    \leq [\boldsymbol{u}_{max}; \boldsymbol{\dot{u}}_{max}]
\end{equation}
where $\boldsymbol{u}_{min}$ and $\boldsymbol{u}_{max}$  
($\boldsymbol{\dot{u}}_{min}$ and $\boldsymbol{\dot{u}}_{max}$)  
are the lower and upper limits of $\boldsymbol{u}$ ($\boldsymbol{\dot{u}}$), respectively.

% ---------------------------------------------------------------------------------------------- %
\subsubsection{QP 1}
Both RCM constraint and visual servoing task are modeled as equality constraints in this QP-based method~\cite{2025huangAccelerated}:
\begin{align}
    \text{min.} \quad & \boldsymbol{\dot{u}}^{T}\boldsymbol{W}_{u}\boldsymbol{\dot{u}} \\
    \text{s.t.} \quad 
    & \boldsymbol{J}_{rcm,i}\boldsymbol{\dot{u}}= \boldsymbol{v}_{rcm,des} \\
    & \boldsymbol{J}_{t,i}\boldsymbol{\dot{u}} = \dot{\boldsymbol{s}}_{des} \\
    &     [\boldsymbol{u}_{min}; \boldsymbol{\dot{u}}_{min}] 
    \leq [\boldsymbol{u}; \boldsymbol{\dot{u}}] 
    \leq [\boldsymbol{u}_{max}; \boldsymbol{\dot{u}}_{max}]
\end{align}
where $\boldsymbol{W}_{u} \in \mathbb{R}^{N_{u} \times N_{u}}$ is a positive definite weight matrix,
and $N_{u}$ is the dimension of $\boldsymbol{\dot{u}}$.

% ---------------------------------------------------------------------------------------------- %
\subsubsection{QP 2}
In this method, the RCM constraint is modeled as an equality constraint, 
while visual servoing task is considered in the cost function~\cite{2025zhouControl}:
\begin{align}
    \text{min.} \quad & \boldsymbol{c}_{s}^{T}\boldsymbol{W}_{s}\boldsymbol{c}_{s} + \boldsymbol{\dot{u}}^{T}\boldsymbol{W}_{u}\boldsymbol{\dot{u}} \\
    \text{s.t.} \quad 
    & \boldsymbol{J}_{rcm,i}\boldsymbol{\dot{u}}= \boldsymbol{v}_{rcm,des} \\
    &     [\boldsymbol{u}_{min}; \boldsymbol{\dot{u}}_{min}] 
    \leq [\boldsymbol{u}; \boldsymbol{\dot{u}}] 
    \leq [\boldsymbol{u}_{max}; \boldsymbol{\dot{u}}_{max}]
\end{align}
where $\boldsymbol{c}_{s} = \boldsymbol{J}_{t,i}\boldsymbol{\dot{u}} - \dot{\boldsymbol{s}}_{des}$
and $\boldsymbol{W}_{s} \in \mathbb{R}^{N_{f} \times N_{f}}$ is a positive definite weight matrix.

% ---------------------------------------------------------------------------------------------- %
\subsubsection{QP 3}
Hierarchical QP with RCM constraint in the first priority and visual servoing task in the second priority are constructed in this QP-based method~\cite{2023colanConstrained}.
The first QP problem is designed to satisfies the RCM constraint:
\begin{align}
    \text{min.} \quad & \boldsymbol{c}_{rcm}^{T}\boldsymbol{W}_{rcm}\boldsymbol{c}_{rcm} + 
    \boldsymbol{\dot{u}}^{T}\boldsymbol{W}_{u}\boldsymbol{\dot{u}} \\
    \text{s.t.} \quad 
    &     [\boldsymbol{u}_{min}; \boldsymbol{\dot{u}}_{min}] 
    \leq [\boldsymbol{u}; \boldsymbol{\dot{u}}] 
    \leq [\boldsymbol{u}_{max}; \boldsymbol{\dot{u}}_{max}]
\end{align}
where $\boldsymbol{c}_{rcm} = \boldsymbol{J}_{rcm,i}\boldsymbol{\dot{u}} - \boldsymbol{v}_{rcm,des}$
and $\boldsymbol{W}_{rcm} \in \mathbb{R}^{N_{rcm} \times N_{rcm}}$ is a positive definite weight matrix.
By solving the first QP problem, the control value $\boldsymbol{\dot{u}}_{rcm}$ is obtained.
Furthermore, the second QP problem is designed to achieve the visual servoing task in the null space of the RCM constraint:
\begin{align}
    \text{min.} \quad &
    \boldsymbol{c}_{s}^{T}\boldsymbol{W}_{s}\boldsymbol{c}_{s} + \boldsymbol{\dot{u}}^{T}\boldsymbol{W}_{u}\boldsymbol{\dot{u}} \\
    \text{s.t.} \quad
    & \boldsymbol{\dot{u}} = \boldsymbol{\dot{u}}_{rcm} + \boldsymbol{N}_{rcm,i}\boldsymbol{\dot{u}}_{t} \\
    &     [\boldsymbol{u}_{min}; \boldsymbol{\dot{u}}_{min}] 
    \leq [\boldsymbol{u}; \boldsymbol{\dot{u}}] 
    \leq [\boldsymbol{u}_{max}; \boldsymbol{\dot{u}}_{max}]
\end{align}

% ---------------------------------------------------------------------------------------------- %
\subsection{IK-based Method}
The IK-based method applies to rigid laparoscopic robots under the assumption that the RCM constraint is perfectly satisfied, allowing a direct relationship between the velocity twists of the connector and the end camera~\cite{2025zhouLaparoscope}.
It first computes the desired end-camera twist from the visual servoing task, 
then determines the corresponding connector twist under the RCM constraint, 
and finally derives the control input via inverse kinematics.

For a rigid laparoscopic robot, the relationship between the connector and end-camera velocity twists is given by:
\begin{equation}
    \begin{bmatrix}
        {}^{b}\boldsymbol{\omega}_{b} \\
        {}^{b}\boldsymbol{v}_{b}
    \end{bmatrix} = 
    \begin{bmatrix}
        \boldsymbol{I}_{3\times3} & \boldsymbol{0}_{3\times3} \\
        [{}^{b}\boldsymbol{p}_{c}]_{\times} & \boldsymbol{I}_{3\times3}
    \end{bmatrix}
    \begin{bmatrix}
        {}^{c}\boldsymbol{\omega}_{c} \\
        {}^{c}\boldsymbol{v}_{c}
    \end{bmatrix}
    \label{eq:vc2vb}
\end{equation}
where ${}^{b}\boldsymbol{p}_{c}=[0;0;L_{rod}]$ is the position of frame $\{c\}$ with respect to frame $\{b\}$,
and ${}^{c}\boldsymbol{\omega}_{c}\in\mathbb{R}^{3}$ and ${}^{c}\boldsymbol{v}_{c}\in\mathbb{R}^{3}$
are the angular and linear velocities of the laparoscope's end camera in frame $\{c\}$, respectively.
Assuming the RCM constraint is perfectly satisfied, 
combining Eq.~(\ref{eq:rcm2_adj}) and Eq.~(\ref{eq:vc2vb}) yields:
\begin{align}
    {}^{rcm}v_{rcm,x} & = (L_{out}-L_{rod}){}^{c}\omega_{c,y} + {}^{c}v_{c,x} = 0 \\
    {}^{rcm}v_{rcm,y} & = (L_{rod}-L_{out}){}^{c}\omega_{c,x} + {}^{c}v_{c,y} = 0
\end{align}
which further lead to:
\begin{align}
    {}^{c}\omega_{c,x} & = -{}^{c}v_{c,y}/(L_{rod}-L_{out}) \\
    {}^{c}\omega_{c,y} & = -{}^{c}v_{c,x}/(L_{out}-L_{rod})
\end{align}
where $\cdot{}_{,x}$ and $\cdot{}_{,y}$ denote the $x$ and $y$ components of the corresponding vector.
Substituting these expressions into Eq.~(\ref{eq:vc2vb}) gives the form commonly used in~\cite{2025zhouLaparoscope}:
\begin{equation}
    \begin{bmatrix}
        {}^{b}\boldsymbol{\omega}_{b} \\
        {}^{b}\boldsymbol{v}_{b}
    \end{bmatrix} = 
    \begin{bmatrix}
        0 & 0 & 0 & 0 & -\frac{1}{L_{in}} & 0 \\
        0 & 0 & 0 & \frac{1}{L_{in}} & 0 & 0 \\
        0 & 0 & 1 & 0 & 0 & 0 \\
        0 & 0 & 0 & -\frac{L_{out}}{L_{in}} & 0 & 0 \\
        0 & 0 & 0 & 0 & -\frac{L_{out}}{L_{in}} & 0 \\
        0 & 0 & 0 & 0 & 0 & 1 \\
    \end{bmatrix}
    \begin{bmatrix}
        {}^{c}\boldsymbol{\omega}_{c} \\
        {}^{c}\boldsymbol{v}_{c}
    \end{bmatrix}
\end{equation}
where $L_{in} = L_{rod} - L_{out}$ denotes the shaft length inside the body.
The control input is then obtained via the robot arm inverse kinematics.

%%%%%%%%%%%%%%%%%%%%%%%%%%%%%%%%%%%%%%%%%%%%%%%%%%%%%%%%%%%%%%%%%%%%%%%%%%%%%%%%%
\section{Simulation Studies}

This section presents simulation studies to evaluate the proposed benchmark across different RCM modeling methods and control frameworks, 
and to analyze the structural insights revealed under controlled conditions. 
The primary results are obtained using a rigid laparoscopic robot, while additional experiments on both rigid and flexible systems are provided in the supplementary video.

% ---------------------------------------------------------------------------------------------- %
\subsection{Simulation Setup}
\begin{figure}
    \centering
    \includegraphics[width=0.4\textwidth]{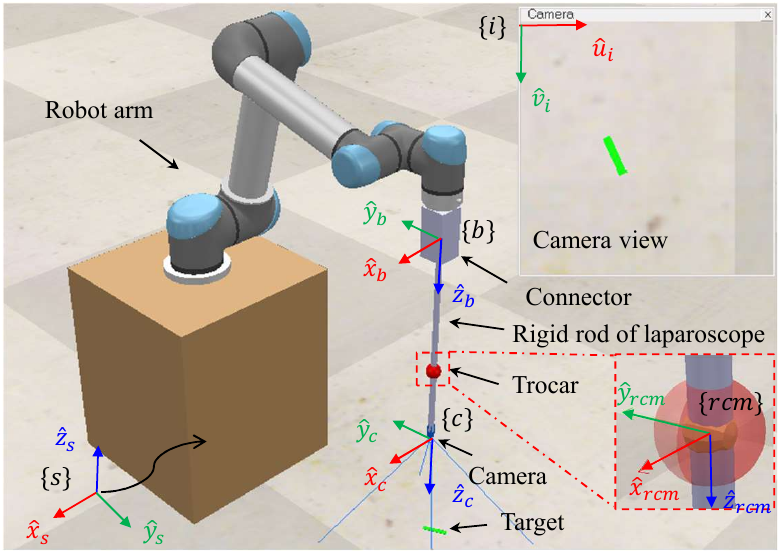} 
    \caption{Simulation setup in CoppeliaSim.}
    \label{fig:simulation_setup}
\end{figure}

A simulation platform is developed in CoppeliaSim to evaluate FoV adjustment, as shown in Fig.~\ref{fig:simulation_setup}.
The system consists of a 6-DoF UR5 manipulator holding a rigid laparoscope. 
A red sphere and a green rectangle represent the trocar and the visual tracking target (e.g., a surgical tool), respectively. 
A dummy attached to the shaft visualizes the RCM point.
The coordinate frames follow Section~II, with frame $\{s\}$ coincident with the CoppeliaSim world frame. 
The image frame $\{i\}$ is defined at the top-left corner of the camera view, 
with the $u$-axis pointing right and the $v$-axis downward. 
The image resolution is $256 \times 256$ pixels.
Controllers are implemented in MATLAB and communicate with CoppeliaSim via the remote API. 
Variables are discretized with time step $\Delta t = 0.1$, 
and joint velocity is computed as $\dot{\boldsymbol{\Theta}}^{k} = (\boldsymbol{\Theta}^{k}-\boldsymbol{\Theta}^{k-1})/\Delta t$, where $k$ denotes the time index.

% ---------------------------------------------------------------------------------------------- %
\subsection{4-DoF Visual Servoing}

Following Huang \textit{et al.}~\cite{2023huang4DOF}, 
the visual feature vector is defined by the position, scale, and orientation of the target rectangle in frame $\{i\}$:
\begin{equation}
    \boldsymbol{s} =[s_1; s_2; s_3; s_4]=[u_c; v_c; Z^{*} \sqrt{a^{*}/a}; \alpha]
\end{equation}
where $[u_c; v_c]$ is the mass center of the target rectangle in the image,
$a$ and $a^{*}$ are the current and desired area of the target rectangle in the image,
and $Z^{*}$ is the desired depth in frame $\{c\}$.
When the area approaches the expected value, $s_3$ approximates $Z^{*}$.
$\alpha = atan2(2\mu_{11}/(\mu_{20}-\mu_{02}))\cdot 90/\pi$ indicates the rotation angle between the long axis of the rectangle and the $u$-axis of frame $\{i\}$,
where $\mu_{11}$, $\mu_{20}$, and $\mu_{02}$ are second-order centered moments computed by
$\mu_{ij} = \sum_{n=1}^{4} (u_n - u_c)^i (v_n - v_c)^j$,
with $[u_n; v_n]$ denoting the $n$-th vertex in frame $\{i\}$.
In the following simulation cases, the desired visual feature is $[128; 128; 100; 70]$.

% ---------------------------------------------------------------------------------------------- %
\subsubsection{Case 1: Tangent Plane Selection in RCM 1}
Since the original work~\cite{2010azimianConstrained} \textit{does not} specify how the tangent plane should be defined, 
this case evaluates how different tangent-plane selections affect the performance of RCM 1.
Two representative choices are compared: planes perpendicular to the $z$-axis of frame $\{s\}$ and frame $\{rcm\}$, respectively. 
The former constrains the RCM point within the $x$–$y$ plane of $\{s\}$, 
corresponding to the implicit setting in~\cite{2010azimianConstrained},
while the latter aligns with the physical interpretation of the trocar constraint and becomes equivalent to RCM 2 with $\boldsymbol{v}_{rcm,des}=\boldsymbol{0}$. 
QP 1 is used as the control framework.

\begin{figure}
    \centering
    \includegraphics[width=0.45\textwidth]{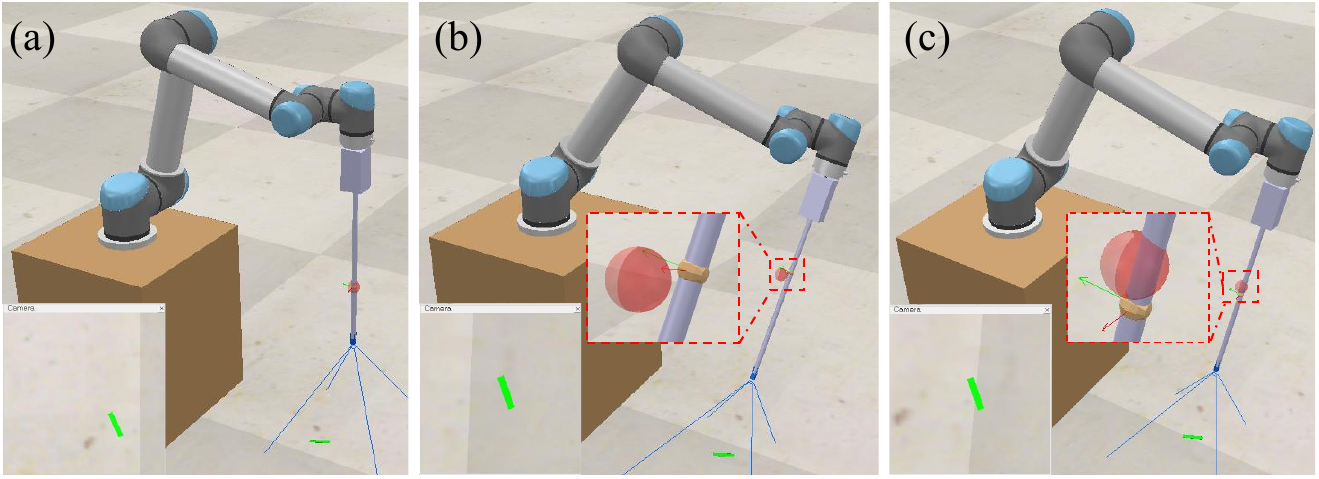} 
    \caption{Snapshots at:
    (a) $t=0$ s in Case 1.
    (b) $t=10$ s in Case 1 when the tangent plane is perpendicular to the $z$-axis of frame $\{s\}$.
    (c) $t=5$ s in Case 2 when using RCM 3 ($N_{rcm}=2$).
    }
    \label{fig:snapshots}
\end{figure}

In this case, the target moves along a straight line:
\begin{equation}
    {}^{s}\boldsymbol{p}_{tar}^{k} = {}^{s}\boldsymbol{p}_{tar}^{1} + ({}^{s}\boldsymbol{p}_{tar}^{K}-{}^{s}\boldsymbol{p}_{tar}^{1}) \times t/T
    \label{eq:line}
\end{equation}
where ${}^{s}\boldsymbol{p}_{tar}^{k}\in \mathbb{R}^3$ is the position of the target at time step $k$ in frame $\{s\}$,
${}^{s}\boldsymbol{p}_{tar}^{1} = [-50;525;5]$ mm and ${}^{s}\boldsymbol{p}_{tar}^{K} = [50;525;5]$ mm
are the initial position and final position of the target in frame $\{s\}$, respectively,
$K=T/\Delta t+1$ is the total time step number,
$T=10$ s is the total simulation time,
and $t=(k-1)\Delta t$ is the current time.
Fig.~\ref{fig:snapshots}(a) shows the initial snapshot of this case,
where the initial visual feature is $[178.09;177.69;139.25;64.98]$.
\begin{figure}
    \centering
    \includegraphics[width=0.45\textwidth]{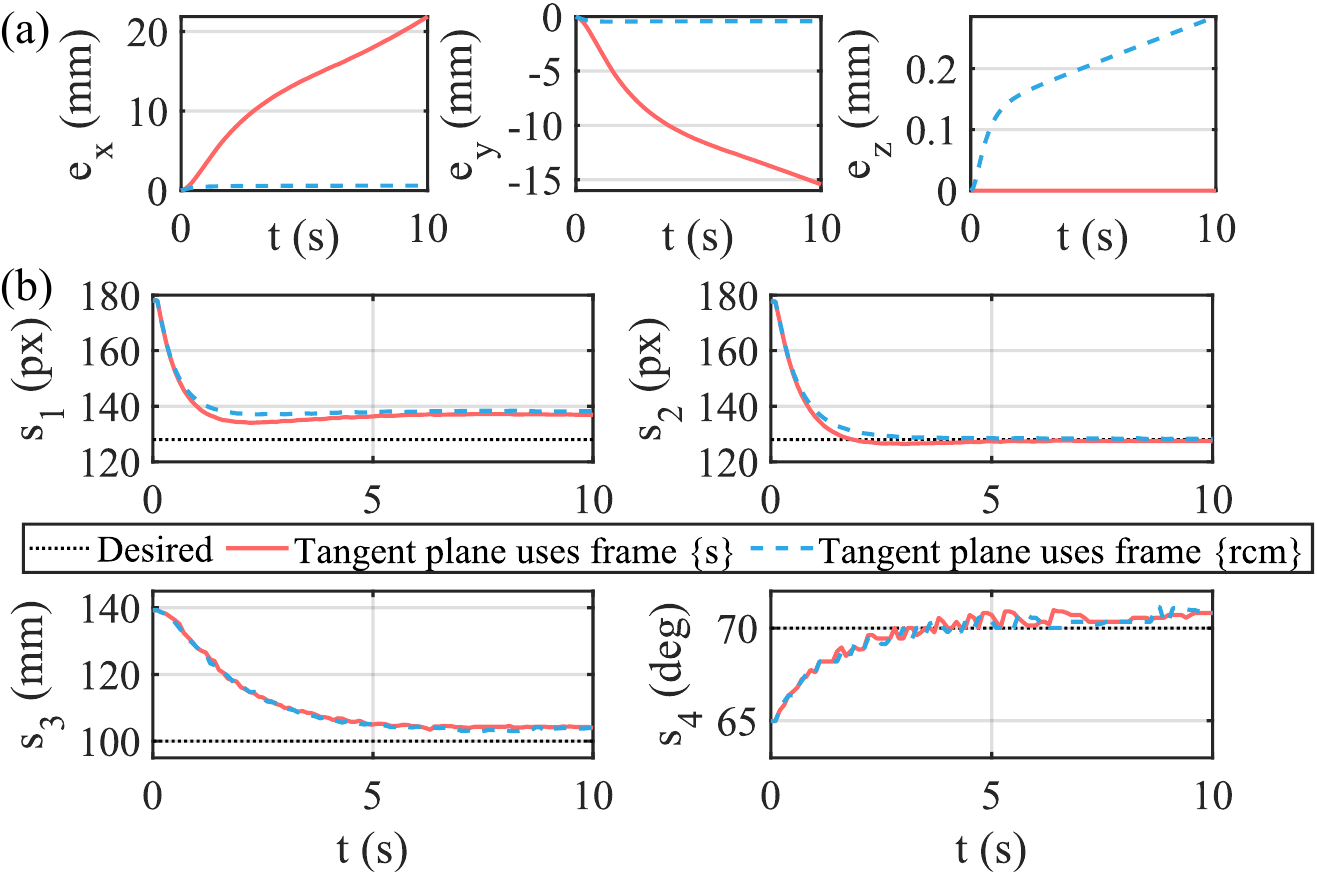} 
    \caption{Simulation results of Case 1, comparing two tangent plane selections in RCM 1.
    (a) RCM errors v.s. time.
    (b) Visual features v.s. time.
    }
    \label{fig:sim1}
\end{figure}

The results are shown in Fig.~\ref{fig:sim1}. 
In both cases, the visual servoing task converges successfully (Fig.~\ref{fig:sim1}(b)). 
However, when the tangent plane is perpendicular to the $z$-axis of $\{s\}$ (red solid), 
significant RCM errors accumulate in the $x$ and $y$ directions (Fig.~\ref{fig:sim1}(a)), 
indicating poor constraint enforcement. 
The deviation is visible in Fig.~\ref{fig:snapshots}(b). 
This occurs because scale adjustment requires insertion of the laparoscope,
which introduces lateral motion of the RCM point in frame $\{s\}$ under this plane definition.
In contrast, defining the tangent plane perpendicular to the $z$-axis of $\{rcm\}$ (blue dash) maintains small RCM errors throughout the motion, 
with maximum errors of $0.63$ mm and $-0.43$ mm in the $x$ and $y$ directions, respectively.

% ---------------------------------------------------------------------------------------------- %
\subsubsection{Case 2: Effects of $N_{rcm}$ for RCM 3}

This case examines the effect of constraint dimensionality in RCM 3 
by comparing $N_{rcm}=2$ (no constraint along the $z$ direction) and $N_{rcm}=3$. 
The same straight-line trajectory in Eq.~(\ref{eq:line}) is used, with QP 1 as the control framework.
\begin{figure}
    \centering
    \includegraphics[width=0.45\textwidth]{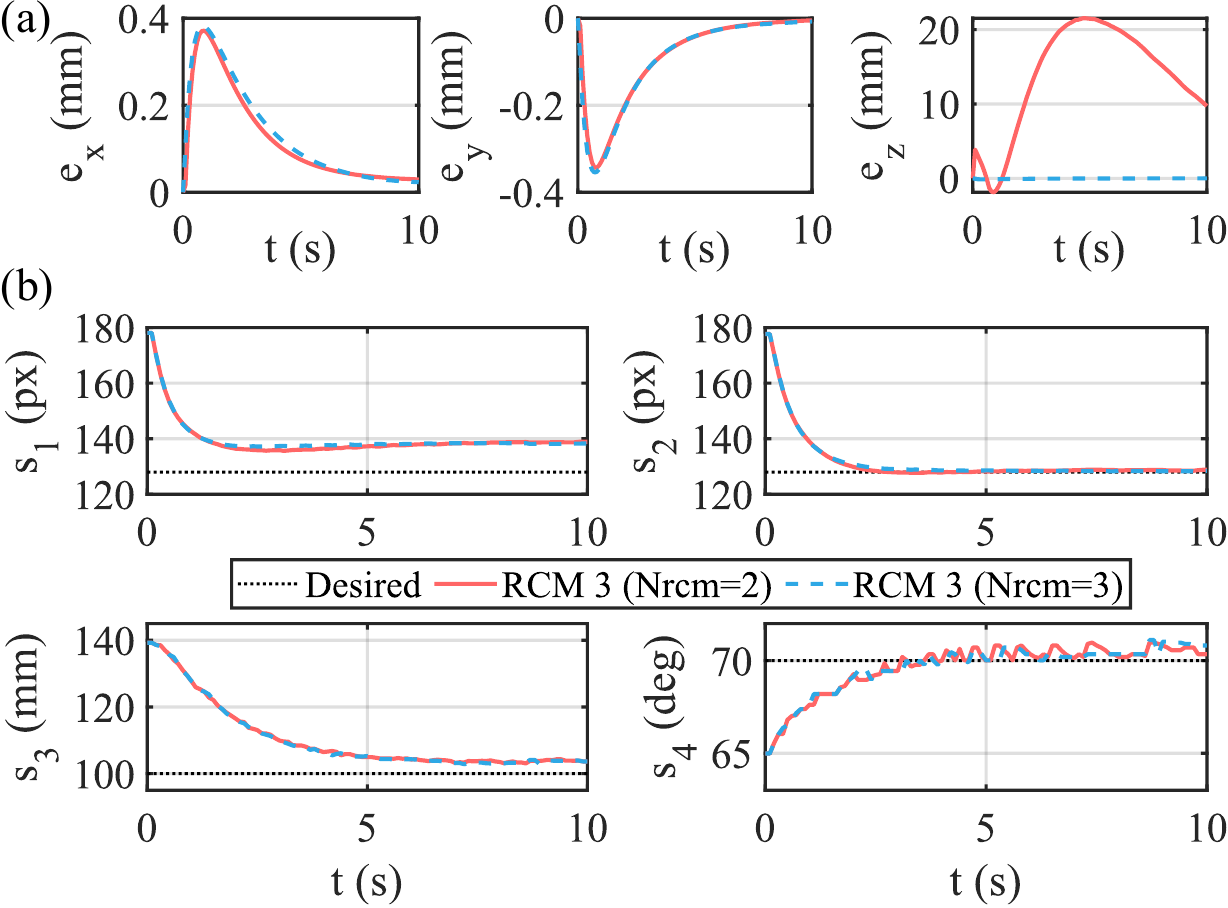} 
    \caption{Simulation results of Case 2, comparing RCM 3 with $N_{rcm}=2$ and $N_{rcm}=3$.
    (a) RCM errors v.s. time.
    (b) Visual features v.s. time.
    }
    \label{fig:sim2}
\end{figure}

As shown in Fig.~\ref{fig:sim2}(b), the visual servoing task converges in both cases. 
However, when $N_{rcm}=2$ (red solid), a large RCM error appears in the $z$ direction, 
while the $x$ and $y$ errors remain small (Fig.~\ref{fig:sim2}(a)). 
In contrast, enforcing $N_{rcm}=3$ (blue dash) maintains small errors in all directions.
The deviation is evident in Fig.~\ref{fig:snapshots}(c), 
where the RCM point shifts along the shaft when $N_{rcm}=2$, 
producing a $z$-direction error exceeding $20$ mm. 
This behavior arises because the virtual variable $\lambda_{3}$ is unconstrained along the axial direction,
leading to drift of the computed RCM location.
These results demonstrate that enforcing all three constraint components ($N_{rcm}=3$) is necessary to ensure correct RCM localization in RCM 3.

% ---------------------------------------------------------------------------------------------- %
\subsubsection{Case 3: Comparison of Different RCM Methods with Open-Loop and Closed-Loop Formulations}
This case compares open-loop and closed-loop RCM enforcement strategies. 
Specifically, RCM 1, RCM 2 with $\boldsymbol{v}_{rcm,des}=\boldsymbol{0}$, 
and RCM 3 with $\boldsymbol{v}_{rcm,des}=\boldsymbol{0}$ (denoted as RCM 2 (0) and RCM 3 (0)) operate in open loop. 
In contrast, RCM 2 and RCM 3 with $\boldsymbol{v}_{rcm,des}$ computed from the RCM position error operate in closed loop. 
All methods are evaluated under the straight-line trajectory in Eq.~(\ref{eq:line}), 
with QP 1 as the control framework.

%Comparison results are shown in Fig.~\ref{fig:sim3}.
%The open-loop methods RCM 1 (red), RCM 2 (0) (blue), and RCM 3 (0) (purple) maintain non-zero RCM errors after the visual servoing task is completed, 
%whereas the closed-loop methods RCM 2 (green) and RCM 3 (orange) effectively drive the RCM errors to nearly zero.

%\RD{SHORTER VERSION: The results are shown in Fig.~\ref{fig:sim3}. 
%The open-loop methods (RCM 1, RCM 2 (0), and RCM 3 (0)) exhibit persistent RCM errors after convergence of the visual servoing task. In contrast, the closed-loop formulations (RCM 2 and RCM 3) drive the RCM error to nearly zero.
%These results highlight the importance of explicit error feedback in enforcing the RCM constraint, as open-loop formulations cannot eliminate residual deviations.}

\begin{figure}
    \centering
    \includegraphics[width=0.45\textwidth]{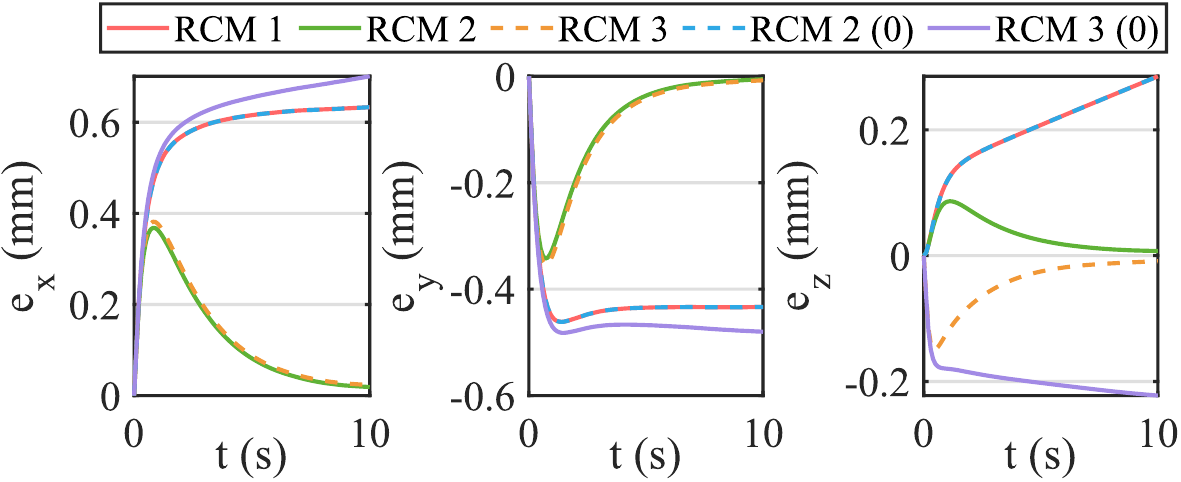} 
    \caption{RCM errors of Case 3, comparing different RCM methods.
    }
    \label{fig:sim3}
\end{figure}

The results are shown in Fig.~\ref{fig:sim3}. 
The open-loop methods (RCM 1 (red solid), RCM 2 (0) (blue dash), and RCM 3 (0) (purple solid)) 
exhibit non-zero steady-state RCM errors after the visual servoing task converges. 
As seen in the $e_x$ and $e_y$ plots, residual lateral errors persist, 
while the $e_z$ plot reveals axial drift, particularly in RCM 3 (0). 
This indicates that purely geometric enforcement without error feedback cannot eliminate accumulated deviation.
In contrast, the closed-loop methods (RCM 2 (green solid) and RCM 3 (yellow dash)) drive the RCM errors toward zero in all directions, 
maintaining sub-millimeter accuracy throughout the motion.
These results highlight the necessity of explicit RCM error feedback to prevent steady-state and axial deviations.

% ---------------------------------------------------------------------------------------------- %
\subsubsection{Case 4: Comparison of Different Control Frameworks}

This case compares the performance of different control frameworks including IK, PI 1, PI 2, QP 1, QP 2, and QP 3. 
IK operates in open loop by enforcing ${}^{rcm}v_{rcm,x}=0$ and ${}^{rcm}v_{rcm,y}=0$, while the remaining five methods use closed-loop RCM 2. 
Two scenarios are considered.

\begin{figure}
    \centering
    \includegraphics[width=0.45\textwidth]{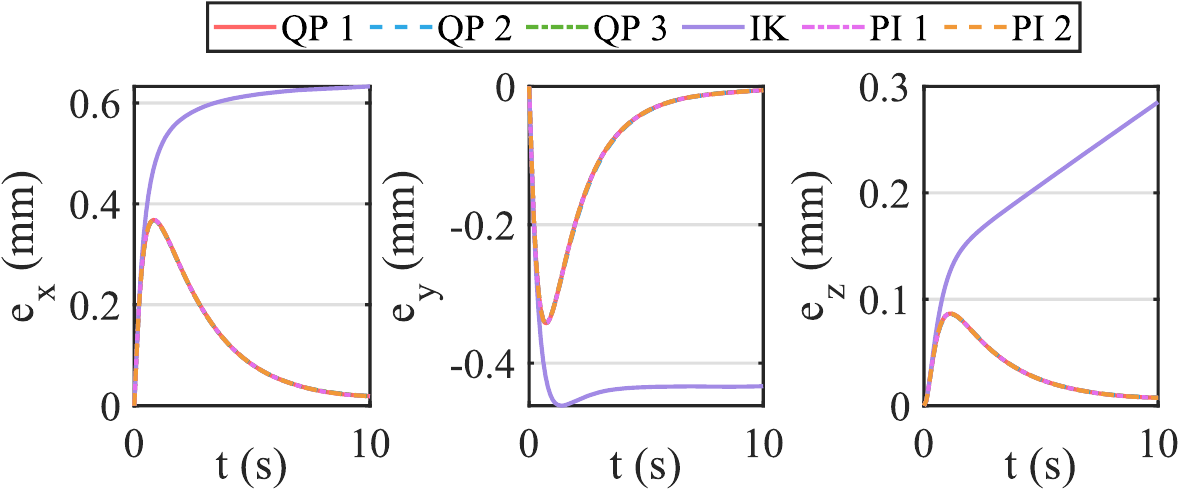} 
    \caption{RCM errors of Case 4 under different control frameworks in the first scenario.}
    \label{fig:sim4_1}
\end{figure}

In the first scenario, the target follows the straight-line trajectory in Eq.~(\ref{eq:line}). 
As shown in Fig.~\ref{fig:sim4_1}, all methods maintain small RCM errors. 
However, IK (purple solid) cannot eliminate steady-state error due to its open-loop formulation. 
The rest closed-loop methods limit the peak RCM error to approximately $0.5$ mm and converge rapidly to near zero.

\begin{figure}
    \centering
    \includegraphics[width=0.48\textwidth]{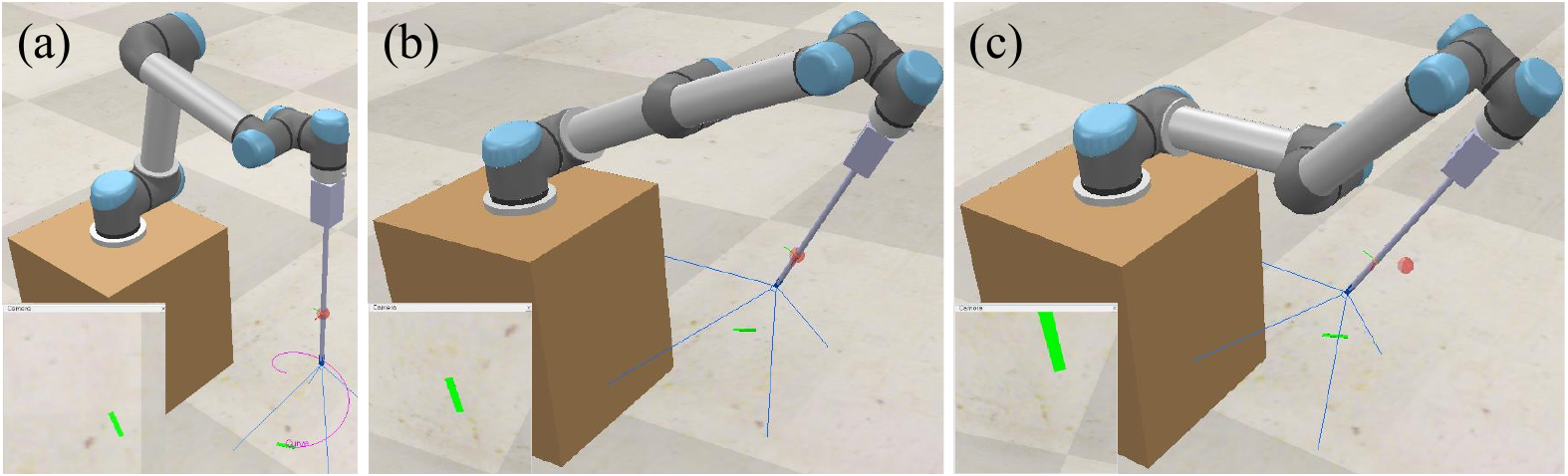} 
    \caption{Snapshots of the simulation in the second scenario of Case 4 using PI 1:
    (a) At $t=0$ s.
    (b) At $t=44.2$ s.
    (c) At $t=48$ s.
    }
    \label{fig:sim4_2_snapshots}
\end{figure}

The second scenario evaluates robustness near singular configurations. 
The target follows a helical trajectory (Fig.~\ref{fig:sim4_2_snapshots}(a)):
\begin{equation}
    {}^{s}\boldsymbol{p}_{tar}^{k} = 
    \begin{bmatrix}
        -150+100\cos(2\pi t/T) \\
        525+100\sin(2\pi t/T) \\
        5+250t/T
    \end{bmatrix}
\end{equation}
The values of the above equation are all in mm, 
and the total simulation time $T$ is set to 60s. 
PI 1, QP 1, and IK are used as the control frameworks in this scenario.

\begin{figure}
    \centering
    \includegraphics[width=0.45\textwidth]{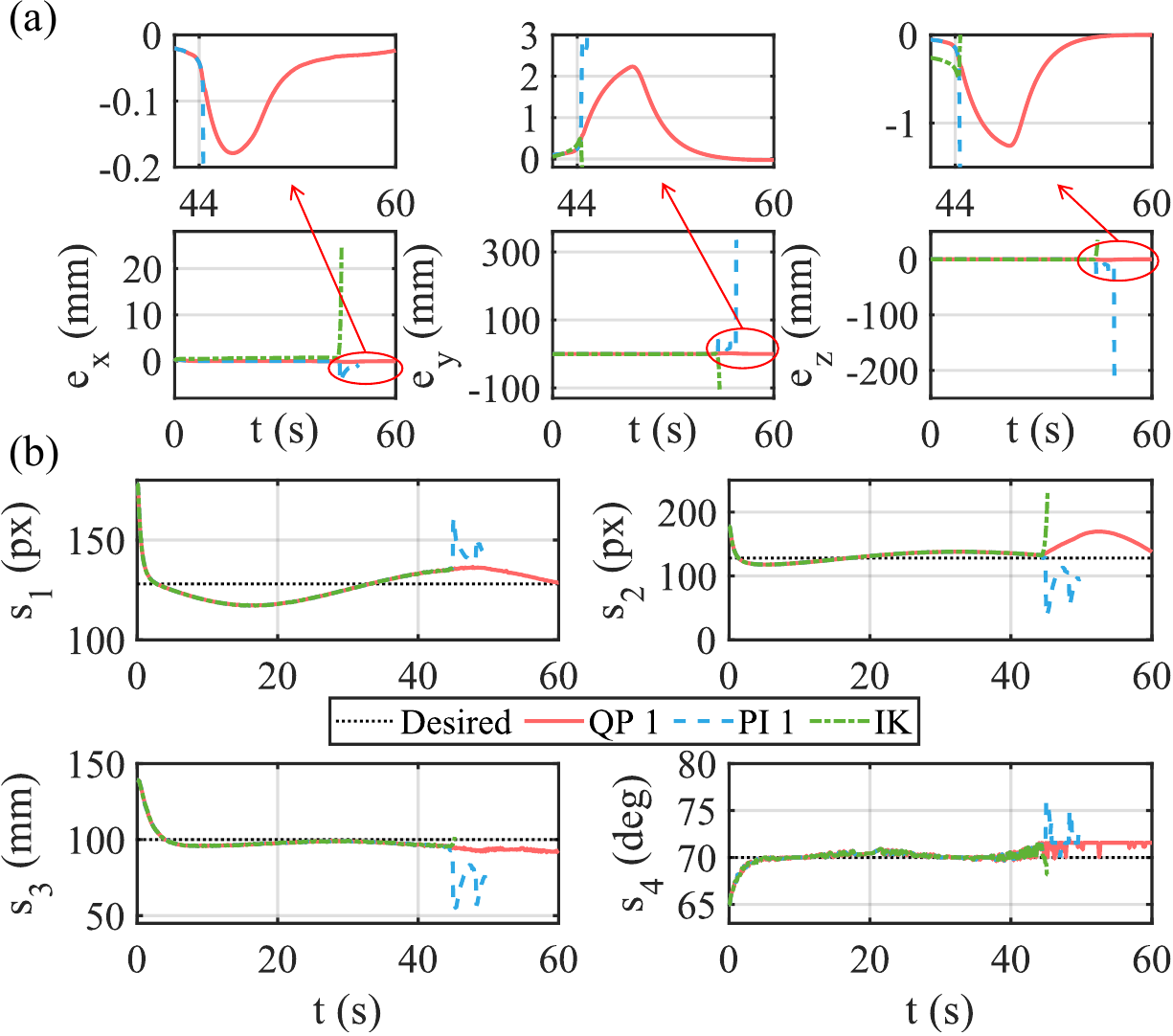} 
    \caption{Simulation results of Case 4 in the second scenario.
    (a) RCM errors v.s. time with zoom-in views.
    (b) Visual features v.s. time.}
    \label{fig:sim4_2}
\end{figure}

As shown in Fig.~\ref{fig:sim4_2}, PI 1 (blue dash) fails shortly after passing through the singular configuration (around $50$ s), 
resulting in large RCM deviation and loss of visual tracking (Fig.~\ref{fig:sim4_2_snapshots}(c)). 
A snapshot at $t=48$ s in Fig.~\ref{fig:sim4_2_snapshots}(c) shows a pronounced deviation of the RCM point from the trocar,
accompanied by a loss of tracking in the visual features.
IK (green dash) exhibits similar failure. 
In contrast, QP 1 (red solid) maintains task convergence and RCM enforcement, 
although transient error fluctuations occur near the singularity (Fig.~\ref{fig:sim4_2}(a)).
These results indicate that QP-based formulations provide improved robustness near singular configurations by explicitly handling task coupling and constraints, 
whereas PI- and IK-based methods are more sensitive to singularity-induced ill-conditioning.


% ---------------------------------------------------------------------------------------------- %
\subsection{Case 5: 2-DoF Visual Servoing}
\begin{figure}
    \centering
    \includegraphics[width=0.48\textwidth]{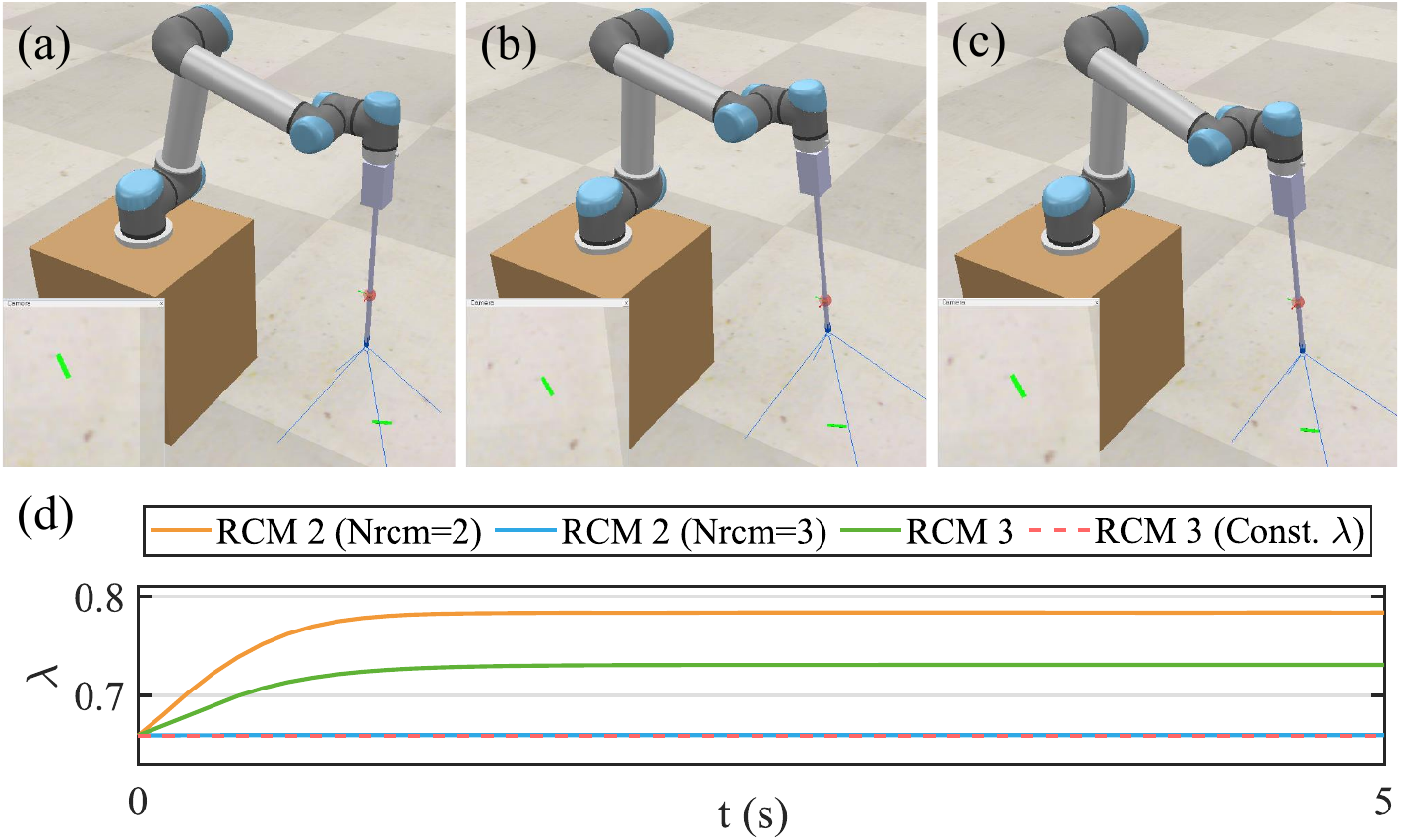} 
    \caption{Simulation results of Case 5, comparing different RCM modeling methods when performing 2-DoF visual servoing.
    (a)-(c) Snapshots at: $t=0$ s;
    $t=5$ s when using RCM 2 ($N_{rcm}=2$);
    $t=5$ s when using RCM 2 ($N_{rcm}=3$).
    (d) $\lambda$ v.s. time.
    }
    \label{fig:sim5_snapshots}
\end{figure}

Some studies consider 2-DoF visual servoing, where only the image position $[u_c; v_c]$ is regulated~\cite{2025caoUncalibrated}. 
Therefore Case 5 compares the performance of different RCM modeling methods and evaluates RCM behavior under this reduced task. 
RCM 2 and RCM 3 are tested using QP 1 as the controller. 
For RCM 2, both $N_{rcm}=2$ and $N_{rcm}=3$ are considered, denoted as RCM 2 ($N_{rcm}=2$) and RCM 2 ($N_{rcm}=3$), respectively. 
For RCM 3, $\lambda$ is either free or fixed at its initial value (denoted as RCM 3 (Const. $\lambda$)) to enforce a constant RCM location.
The target is static at $[-150;625;5]$ mm in frame $\{s\}$, with initial image position $[95.83;95.35]$ pixel. 
The simulation duration is $T=5$ s.

\begin{figure}
    \centering
    \includegraphics[width=0.45\textwidth]{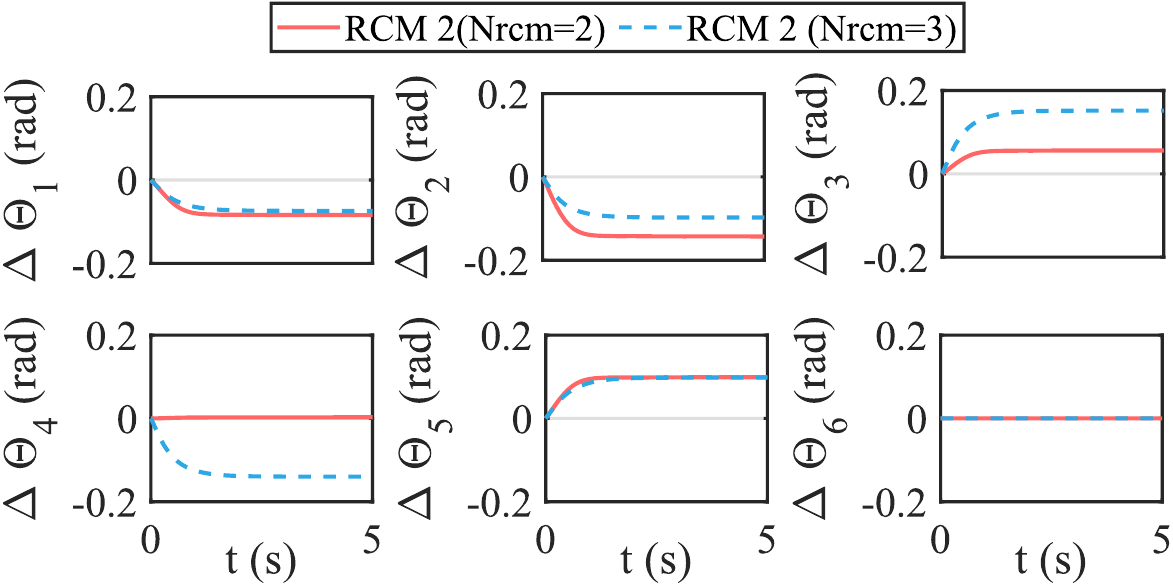} 
    \caption{The changes of joint angles in Case 5 when using RCM 2 with $N_{rcm}=2$ and $N_{rcm}=3$.
    }
    \label{fig:sim5_joint}
\end{figure}

Snapshots for RCM 2 ($N_{rcm}=2$) and RCM 2 ($N_{rcm}=3$) are shown in Fig.~\ref{fig:sim5_snapshots}(a)-(c). 
Fig.~\ref{fig:sim5_snapshots}(a) presents the initial configuration at $t=0$ s, 
while Figs.~\ref{fig:sim5_snapshots}(b) and (c) show the final states at $t=5$ s for RCM 2 ($N_{rcm}=2$) and RCM 2 ($N_{rcm}=3$), respectively. 
The corresponding joint-angle variations are provided in Fig.~\ref{fig:sim5_joint}, 
and the evolution of $\lambda$ is shown in Fig.~\ref{fig:sim5_snapshots}(d).
When RCM 2 ($N_{rcm}=3$) is used (blue solid), $\lambda$ remains constant at $0.66$ throughout the FoV adjustment, 
as expected for a fixed RCM point. 
In contrast, with RCM 2 ($N_{rcm}=2$) (orange solid), $\lambda$ increases from $0.66$ to $0.78$, 
indicating axial retraction of the laparoscope during alignment, as observed in Fig.~\ref{fig:sim5_snapshots}(b).
The configuration cost $\Delta \boldsymbol{\Theta}^{T}\Delta \boldsymbol{\Theta}$ is $0.04$ for RCM 2 ($N_{rcm}=2$), compared to $0.07$ for RCM 2 ($N_{rcm}=3$), 
indicating reduced joint motion when the axial constraint is relaxed. 
A similar increase in $\lambda$ is observed for RCM 3 (green solid).
If a fixed trocar pivot is required, additional constraints on $\lambda$ are necessary to prevent axial drift.

%%%%%%%%%%%%%%%%%%%%%%%%%%%%%%%%%%%%%%%%%%%%%%%%%%%%%%%%%%%%%%%%%%%%%%%%%%%%%%%%%
\section{Discussions}

The proposed playground enables systematic analysis of how RCM modeling choices interact with controller architectures in automatic FoV adjustment. The results reveal several structural insights:
\begin{enumerate}	
	\item \textbf{Impact of RCM modeling choices}:
	RCM formulations differ not only geometrically but also numerically. 
    Tangent-plane-based models (RCM 1) are sensitive to plane definition: 
    defining the plane in frame $\{s\}$ introduces lateral drift during insertion, 
    whereas defining it in the local RCM frame better preserves the physical trocar constraint. In variable-based models (RCM 3), 
    relaxing axial constraint ($N_{rcm}=2$) allows drift of the virtual variable, 
    leading to incorrect RCM localization. These effects stem from how each formulation structures the RCM Jacobian.
	
	\item \textbf{Open-loop versus closed-loop enforcement}:
	Open-loop formulations enforce geometric consistency but lack error feedback, 
    resulting in steady-state residual errors. 
    Closed-loop formulations incorporate RCM position feedback, 
    enabling convergence of constraint violations to zero and preventing accumulated drift.
	
	\item \textbf{Controller robustness and singularity sensitivity}:
	Under nominal motion, PI-, IK-, and QP-based methods perform similarly. 
    However, near kinematic singularities, PI- and IK-based methods become sensitive to Jacobian ill-conditioning. 
    Since pseudoinverse-based updates amplify small singular values, 
    constraint coupling can cause rapid error growth and loss of tracking. 
    In contrast, QP-based controllers solve a constrained optimization problem with explicit regularization and inequality handling, 
    improving numerical robustness and stability near singular configurations.
	
	\item \textbf{Modeling-controller interaction}:
	The overall robustness depends on the interaction between modeling formulation and controller structure. 
    Modeling choices determine the conditioning and dimensionality of the constraint Jacobian, 
    which in turn affects controller sensitivity. 
    Thus, modeling and control cannot be selected independently and their combined structure governs accuracy, stability, and efficiency.
	
	\item \textbf{Design tradeoffs}:
	Relaxing axial constraints reduces joint motion but introduces RCM drift, 
    revealing a tradeoff between strict constraint enforcement and motion efficiency. 
    Controller selection further influences robustness under challenging trajectories.
\end{enumerate}

Overall, the results indicate that robustness in RCM-constrained visual servoing is governed by the structural coupling between modeling formulation and controller architecture.

%%%%%%%%%%%%%%%%%%%%%%%%%%%%%%%%%%%%%%%%%%%%%%%%%%%%%%%%%%%%%%%%%%%%%%%%%%%%%%%%%
\section{Conclusions}
This paper presented a unified benchmark framework integrating three representative RCM modeling approaches and six control architectures for automatic FoV adjustment in laparoscopic robots. 
The comparative study revealed key structural sensitivities in RCM-constrained visual servoing, including the impact of tangent-plane definition, constraint dimensionality, open- versus closed-loop enforcement, and controller robustness near kinematic singularities.
The results demonstrate that modeling formulation and controller design are tightly coupled, and their interaction fundamentally determines accuracy, stability, and robustness. 
Future work will incorporate dynamic effects and evaluate performance under more realistic physical conditions.

\section*{Acknowledgments}
The open-source code of the proposed benchmark is available at: \url{https://github.com/zhangj726/RCM-Constrained-IBVS}.
The authors gratefully acknowledge the assistance of GPT in performing grammar correction and language refinement.
%%%%%%%%%%%%%%%%%%%%%%%%%%%%%%%%%%%%%%%%%%%%%%%%%%%%%%%%%%%%%%%%%%%%%%%%%%%%%%

% \clearpag
\bibliographystyle{IEEEtran}
\bibliography{reference}{}  % .bib
\end{document}